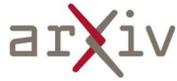

# The Media Inequality, Uncanny Mountain, and the Singularity is Far from Near: Iwaa and Sophia Robot versus a Real Human Being

Johan F. Hoorn *

The Hong Kong Polytechnic University, School of Design, Department of Computing, Hung Hom, Hong Kong SAR, China, johan.f.hoorn [] polyu.edu.hk

Ivy S. Huang

The Hong Kong Polytechnic University, School of Design, Hung Hom, Hong Kong SAR, China, ivyshiming.huang [] connect.polyu.hk

Design of Artificial Intelligence and robotics habitually assumes that adding more humanlike features improves the user experience, mainly kept in check by suspicion of uncanny effects. Three strands of theorizing are brought together for the first time and empirically put to the test: Media Equation (and in its wake, Computers Are Social Actors), Uncanny Valley theory, and as an extreme of human-likeness assumptions, the Singularity. We measured the user experience of real-life visitors of a number of seminars who were checked in either by Smart Dynamics' Iwaa, Hanson's Sophia robot, Sophia's on-screen avatar, or a human assistant. Results showed that human-likeness was not in appearance or behavior but in attributed qualities of being alive. Media Equation, Singularity, and Uncanny hypotheses were not confirmed. We discuss the imprecision in theorizing about human-likeness and rather opt for machines that 'function adequately.'

## 1 INTRODUCTION

"I shall endeavor to function adequately, sir." is what android Data promises commander Ryker in min. 55:50 of *Star Trek: The Next Generation, Encounter at Farpoint* (Allen, Fontana, & Roddenberry, 1987) [1]. Android Data is a robot who gladly gives up its superiority in order to be human whereas that very aspiration does trouble the commander.

Android Data has Artificial Intelligence (AI) installed that reaches far beyond current state-of-the-art systems. However, if we equip a robot with current *natural* intelligence, would not that be a machine that outperforms all stota AI, which is not capable of handling exceptional or 'edge' cases vide the numerous self-driving car accidents lately (e.g., Uber, March 19, 2018) (Li, 2019) [23]? Would not a social robot in remote control do better than an AI-driven machine as its human operator can interpret a situation, have perfect language understanding, and knows politeness rules according to one's culture? Should not a robot in remote control pass the Turing test at least at a higher level than any AI-driven machine? Or would that scare people (cf. commander Ryker's response); a robot that is as good (or bad) as any human being? Or would we celebrate living in a human-computer symbiosis, the old dream of Joseph C. R. Licklider (1960) [24] and of Gill (1996) [7]? Or 'when humans transcend biology,' would we hail Kurzweil's (2005) [20] singularity; or would we dread cyborgization as mentioned by Pelegrin-Borondo, Arias-Oliva, Murata, and Souto-Romero (2018) [39]?

---

* Corresponding author.

> Our sole responsibility is to produce something smarter than we are; any problems beyond that are not *ours* to solve …
> (Kurzweil, 2005, Chap. 2) [19]

### 1.1 Related work: Media Equation

In the current development of interactive robots and embodied agents, human-likeness is one of the strongest guiding principles: Agencies have eyes, they greet, say their name, and more-or-less follow behavioral rules and social scripts. From a designers' and developers' viewpoint, an increase in behavioral realism and appearance would increment user involvement and assumingly stimulate lifestyle changes, better learning results, more hospitality, and higher entertainment value (for an overview of current robotics, consult Henschel, Laban, and Cross (2021) [10]). It may even go so far that claims are made to emulation of human qualities, superintelligence, or 'robot awakenings,' artificial systems becoming sentient and conscious like humans are (e.g., Hayles, 1999 [9]; for a discussion and critique, see Yue, 2022) [49].

In the scientific community as well, human-likeness takes center-stage in studying the effects that robots and embodied agents have on people such as the attribution of emotions and feeling empathy for a machine. One of the earliest examples of theorizing about machines being treated as if they were humans is Media Equation theory (Reeves & Nass, 1996) [40]. Please note the *as if,* because that leaves room for machines not being treated quite like humans. Media Equation and in its wake, Computers Are Social Actors (CASA, Nass & Moon, 2000) [38] have been predominant in studies on human-likeness of machines, stating that people engage with technology in a human-like fashion, applying social scripts even while being consciously aware that what they deal with is nonhuman media and machinery.

> …the more machines present human-like characteristics in a consistent manner, the more likely they are to invoke a social response. (Moshkina, Trickett, & Trafton, 2014) [36]

In this line of thought, designers have been adding humanlike characteristics to robots in looks and behaviors (enhancing 'realism' - Tinwell, Nabi, & Charlton, 2013 [44]; MacDorman, Green, Ho, Koch, 2009 [29]) to improve the user experience (i.e. 'involvement'), examples being Chen Xiaoping's Jia Jia, Hiroshi Ishiguro's Erica, or David Hanson's Sophia.

Gambino, Fox, and Ratan (2020) [6] provide a valuable reconsideration of Media Equation and CASA hypothesizing, pointing out that technology and users have changed – with different interaction schemas and expectations than before. These authors point out two boundary conditions of, in particular, CASA: To evoke a social response, first, machines should offer social cues but it is unclear how many and probably more importantly, it is what users perceive in or attribute to a machine or other objects that makes up the larger part of their responses. Second, agencies should show a degree of (attributed) autonomy, being perceived as the source of communication (cf. a human being) not just transmitting it (cf. a telephone).

To date, researchers indebted to Media Equation (ME) and CASA (e.g., De Graaf, Ben Allouch & Van Dijk, 2015 [3]; Menne & Schwab, 2018, p. 201 [32]) hardly doubt its tenet of humans treating machines as if they were another human, assuming that machines resembling a human would render the ultimate user experience in terms of interaction and relationship. This apparent 'fact' has become encyclopedic (Lee-Won, Joo, & Park, 2020) [22] and over the course of its existence has not been challenged much – except for Bartneck, Rosalia, Menges, and Deckers (2005) [2], who, in countering Media Equation, found that people had fewer concerns to abuse robots than humans.

In quite a few places, the original ME and CASA theorists pointed out that they did not mean to *equate* human responses to socially behaving machines with responses to humans – although the name Media Equation certainly gives that impression.

… individuals can be induced to make attributions toward computers as if the computers were autonomous sources. (Nass & Steuer (1993) [37], p. 511)

Computers … are *close enough* to human that they encourage *social* responses. The encouragement … need not be much. As long as there are some behaviors that suggest a social presence, people will respond accordingly. (Reeves & Nass, 1996 [40], p. 22)

… any medium close enough will get human treatment, even though people know it's foolish and even though they likely deny it afterwards. (Reeves & Nass, 1996 [40], p. 22)

… just as if the machine were a real person with real feelings. (Reeves & Nass, 1996 [40], p. 23)

… individuals are responding mindlessly to computers to the extent that they apply social scripts … (Nass & Moon [38], p. 83)

True as this may be, the 'close enough,' 'as if,' 'to the extent that,' and 'knowing it's foolish' tell that the original ME and CASA authors kept in mind that users may have had some rationalization and reflection about the non-human nature of their social machines, although quite some users may have shown behaviors that *resembled* (not equaled) responding to a human. Thus, 'media equation' is not equation and 'computers are social actors' actually means that computers may be *perceived* as such and may be treated that way *to a certain degree*.

Wykowska (2020) [48] cites a series of investigations, including brain research, confirming human adaptability in using social scripts for entities that are not humans but at least have some resemblance to it (e.g., iCub robot). Hortensius, Hekele, and Cross (2018) [12] also review findings that humans to a degree can identify emotions from robot's facial expressions, using the same brain regions as they do for humans – more-or-less: "… [brain] regions showed attenuated responses to artificial compared to human agents." Vide Wykowska's overview [48], they are *adapted* social scripts, not duplicated social scripts. Given Hortensius, Hekele, and Cross (2018) [12], people interpret the surface representations of a robot's face but the activation level of the same brain regions is lower.

The imprecision pointed out by Gambino, Fox, and Ratan (2020) [6] is crucial to evaluate the reach of human-likeness assumptions. Same brain region would underscore ME and CASA, lower activation maybe not. Or then again, it does, because the robot was not up to par with actual face-to-face human communication. Many predictions fall into the range of observed results as long as humans more-or-less act similarly in interaction with real human beings. Question is, how much similarity or resemblance is still in range of the ME and CASA predictions? For example, a robot making beeping noises may evoke empathy (see Hortensius, Hekele, & Cross, 2018 [12]). Making sounds that resemble vocalization may count as a social cue. However, a human making beeping noises may not evoke empathy but maybe is perceived as humorous. Is human-likeness underscored or not? The social script for humor is different from that of empathy, yet, both are 'social scripts.' The robot was seen as 'different' yet 'cute.'

Apparently, there is a lower threshold after which people start to apply social scripts. From the original publications, one may think that ME and CASA focus on the application of social scripts to media and machines without much concern about the user experience. However, this is not so. In their famous experiment with a television set playing news content (condition 1) versus a TV set with the same information mixed with entertainment (condition 2), Nass and Moon (2000 [38], p. 96) found that people in condition 1 thought the TVs were specialists (i.e. more serious, informative, of higher quality); in condition 2, the TV was regarded a generalist (i.e. less informative, low quality). The authors concluded that people tend to focus on social cues and not on the not-so-social cues. Social cues would evoke social expectations and conventions. Hence, people would apply simple social scripts for interaction.

Differentiation in user responses (serious, informative, quality) was the conclusive ground for Nass and Moon to infer (not measure) that users had applied social scripts to non-social entities like TV sets. The assumption would be that given a minimum set of social cues, social scripts are activated, giving rise to various user experiences (which in the case of a displeasing social cue also may be negative). It is reasonable to infer, then, that with more social cues, the application of social scripts becomes more intense, so that user experiences increase (for good or for ill).

Gambino, Fox, and Ratan (2020 [6], p. 75) discuss a range of CASA-inspired studies, including Nass and colleagues, where, for example, pairing car voice emotion with driver emotion resulted into safer driving behavior, interpersonal disclosure of emotions to chat bots rendered positive effects, and verbal robots were more effective when applying human persuasive strategies. Later authors also confirm that the underlying assumption is that increased human-likeness improves the user experience. In studying what it is for a robot to be 'social,' De Graaf et al. [3] observed that participants frequently mentioned that robots should have "feelings and thoughts," sensing "our presence and moods," inducing a 'feeling of companionship and coziness;' all of which are user-related experiences. The application of social scripts even may directly involve the user experience, for instance, in how to respond to a robot's distress (cf. Menne et al. [32]). Showing empathy to a robot 'in pain' may count as an affective response and so the application of a social script such as offering help is directly related to the user experience as the script is applied with the rise of the user's emotional involvement with the robot. Whether *ex-consequence* or not (Hoorn & Tuinhof, 2022 [11]), measuring user-experience dimensions is what makes ME and CASA researchers infer that people apply social scripts to machines.

Apparently, there is an increase in positive responses assumed when the number of (perceived) social cues increases. Minimally then, an empirical investigation into ME or CASA should include a human-human condition (as proposed by Wykowska, 2020 [48]) and predict an increase in positive experience (of whatever kind) for machines that (are perceived to) provide more or better social cues. Results would be at odds with ME and CASA if at certain tasks, the human would be less appreciated than the machine. Or would there be an optimum? If so, ME and CASA become virtually untestable because any result would fit the paradigm: When confirming human-human behavior, the machine is still in range with optimal human-likeness; when disconfirming human-human behavior, the machine was below threshold or went beyond the optimum of human-likeness.

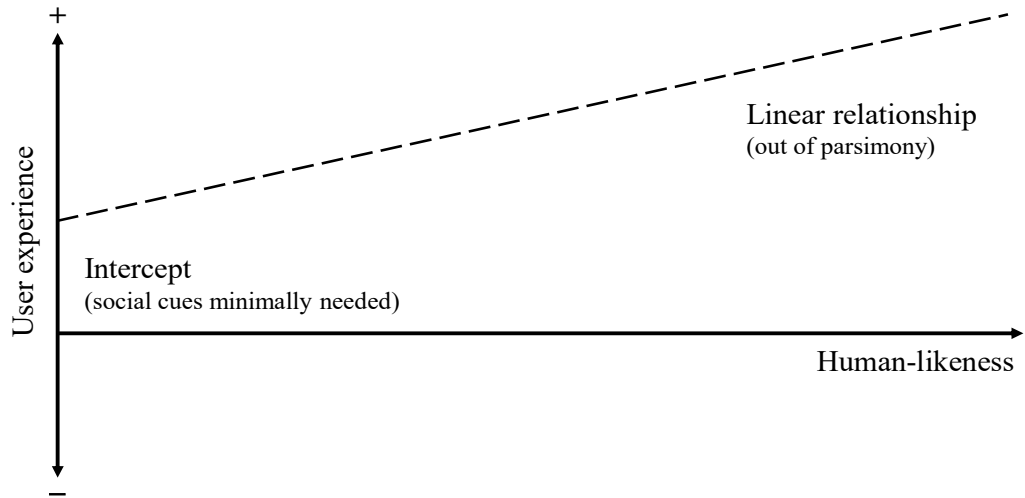

Figure 1: Human-likeness hypotheses expect positive user experience to grow with increased human-likeness (linear conception).

So far, hypotheses of human-likeness do not have clear test predictions. The results we should find would have a lower threshold (i.e. an intercept), an increase in positive responses as a function of (perceived) human-likeness (i.e. a linear relationship), and perhaps an optimum (i.e. an asymptote). Asymptotic growth would be non-linear, but before turning into asymptote, would the expected curve be parabolic, quadratic, exponential, logarithmic, or stepwise even, and could steps be unequal? With theory being mute about the type of relationship between human-likeness and internal response, simple linearity would be the most straightforward and parsimonious choice (Figure 1). That would turn our first hypothesis (H1) into a simple *Human-likeness: the more the better.* This may not be what the original ME and CASA theorists had in mind but they did not explicate what relationship we should be looking for. In the design of robots and AI, H1 is often implicitly assumed.

### 1.2 Technological singularity

Particularly for certain members in the AI community, human emulation or even superseding human abilities is a long-standing goal. The very terminology of neural networks, deep learning, and genetic algorithms already points into that direction but in an extrapolation of H1 (*Human-likeness: the more the better*), certain authors claim that there will be an outburst of recursive intelligence and machine consciousness where boundaries between human and machine existence are blurred, something Vinge (1993 [46]) called 'the singularity.' For an overview of singularity ideas, see Eden, Steinhart, Pearce, and Moor (2012 [4])

These are visions that to a certain degree may bear some validity as people already wear cochlear implants, pacemakers, have artificial dentures, use dialysis machines, have bionic implants, and put microchips under their skin. Superhuman ultra-intelligence (Vinge, 1993 [46]; Kurzweil, 2016 [20]), however, is different from ME and CASA assumptions in that more human-likeness not only grows towards the human standard; it goes beyond it.
5

Yet, as argued, most qualifications of human-likeness (or beyond) are perceptual. Maybe a lonely older adult with little technological experience regards a simple Nao machine as a technological miracle; after all, it talks, walks, and plays a better game of chess. Maybe the AI professor remains a skeptic no matter a system's claimed "potential for the general problem solving ability commonly recognized as true human intelligence" (Williams, 2020 [47]). After all, psychologists already have a hard time demarcating what intelligence actually is: "... the study of it is pervaded by neglect and misconceptions" (Falck, 2020 [5]).

In favor of singularity thinking is that at least a direction is provided in what to look for: the growth of human-likeness (i.e. intelligence) of technology would be 'exponential' (e.g., Growiec, 2022 [8]), which some would welcome (positive user experience) and others would dread (negative user experience). Exponential growth is a kind of power law that is scale-invariant – a re-scaling of the axes would not change the shape (behavior) of the curve. If so, the behavior (not the rate) of the evolution does not depend on the size of the evolutionary system: The trend (i.e. the reproduction rate per entity) of the increase would be the same. Therefore, an exponential function may still be in place even if "… careful consideration of the pace of technology shows that the rate of progress is not constant" (Kurzweil, 2004 [18], p. 382).

It appears that 'exponential growth' is an extrapolation of Moore's Law, which is concerned with the history of baking hardware chips and not with the future of intelligent systems. Moore (1965 [33]) saw a historical trend of an increasing density of microlithographic transistors in integrated semiconductor circuits, which later was connected with an increased number of data bytes processed across the Internet. Yet, past performance is no guarantee of future results. Smaller yet stronger, this is what inspired the singularity's super-linear idea of 'acceleration without end.'

An exponent shows an increase in proportion to its own size, for instance, in the series: 2, 4, 8, 16, 32, etc. However, why would the growth of an AI be exponential only? Why would it not be squared, cubed, or even faster than exponential, for example, a factorial function of *n!*. Is the expected 'exponent' always the same or can it vary over time?

If the exponent does not vary, then what is its standard value? Can we have fractional exponents? What if the sequence is 0.0000002, 0.0000004, 0.0000008, etc.? In that case, the growth curve is almost flat, indicating extremely long incubation before acceleration. Can the value of the exponent also be negative? In that case, Artificial General Intelligence would be on a downward slope.

Kurzweil (2004 [18], p. 394) states that "… the resources deployed for computation are also growing exponentially." Does that mean that there are no influences from the outside world that factor in on the speed of progress? Once beyond human performance, will the break-off point perhaps initialize asymptotic growth? Are not there physical limits such as energy depletion? If not, from where will technology derive energy and materials, the demand of which will be ever-increasing as well? Do technological systems never disintegrate? Does the second law of thermodynamics not apply to exponentially growing systems? Because if it does, technological growth will not be exponential anymore.

Singularity's functional description does not span a limited range of validity. How much deviation from the projected curve is still in range with its predictions? The assumption is that the reproduction rate per entity is a constant (no other interacting factors seem to exist); there seems no limit to the consumption of resources. If at any stage, other factors interact, the trend is unlikely to be exponential. Singularity ignores increased barriers to reproduce marginally more quantity, saturation effects, nonlinear effects of information density, etc. Singularity ignores that near-threshold, phase transition might occur, indicating different behaviors of the system, or even the nature of technology itself, demanding a new set of equations. In a first-order phase transition, the state variables change abruptly; in a second-order phase transition, they change gradually. Therefore, near the boundary, the original equation may become invalid.

"There's even exponential growth in the rate of exponential growth" (Kurzweil, 2004 [18]). As a result of the law of accelerating returns, a double exponential growth is predicted of information-based technology, which, we suppose, would include robots as well. The emphasis is on neural networks, which would "ultimately dominate" computing and emulate the human brain (Kurzweil, 2004 [18], p. 392). The only direction Kurzweil predicts is continuation, claiming that his curve 'matches the available data' but his data set is not available for scrutiny and a goodness-of-fit is not offered.

To make the exponent work for his example, Kurzweil (2004 [18], p. 392) needs a fixed base such as time so that we can estimate the growth rate of neural-network technology per year. To establish that fixed base, Kurzweil needs to do a number of assumptions, neither of which are doubted, investigated, or corroborated by the author; they are personal hunches about costs and about the brain. Neural-network calculations would be cheaper than conventional computation but we already see how deep-learning systems take in huge amounts of data from data farms, which are then processed for days or weeks and return many misclassifications if unsupervised by humans. Nevertheless, neural networks would be "at least" a hundred times cheaper (maybe more, maybe less?) than conventional computation. The factor '100' then would translate into the fixed base Kurzweil needs for showing the exponent, which is 6 years, but these 6 years would be 'approximate' (how much deviation is still acceptable?) and later this century less than 6 years by which the very basis of growth being exponential breaks down as there would be a disruption in the continuation of the curve.

All of this would particularly count for digital control of analog electronics (cf. robots) because that would parallel "the brain's digital controlled analog electrochemical processes" (Kurzweil, 2004 [18], p. 392). This is the human-likeness part of singularity thinking. Yet, few neuroscientists would agree that a brain's control is digital (i.e. binary), and that analog electronics resemble the biochemistry of the brain, communicating not just through electrons but through hormones as well. Nerve cells are of different type and they contain DNA and proteins while every type of nerve cell uses a specific range of genes to build the molecular mechanism for its function. So far, neural networks are not even close to the sophistication of human information processing. According to Kurzweil, the brain would have 100 billion neurons and each neuron would have a thousand connections, where the 'calculations' would primarily take place, which would be 200 per second. However, that 100 billion neurons may be higher or lower. We also do not know if neurons should be our focus (maybe just cell bodies, or synapses, ion shafts, or neurons plus glial cells). Neurons change their connections on a daily basis (so why a 1000?). Neurons do not calculate, they have no brain of their own. Nor do we know the firing rate per second of each brain cell as this changes per stimulus type being processed. Kurzweil (2004 [18], p. 392) admits that he works from little but estimates, which can be higher or lower, but that would shift his projections only "by a relatively small number of years" (ibid.). Again, 'relatively small' leaves room for interpretation and by anticipating a change in the 6-years period later this century, the curve is not a continued exponential function anymore, which is what Kurzweil wished to demonstrate in the first place.

It is laudable that Kurzweil comes up with a calculus to specify singularity predictions. At least, there is something specific to test. His dependent variable is bytes per second for money spent (bps/$), expecting to have a human-brain-like computer by this very year, in 2023 (Kurzweil, 2004 [18], p. 393). The formula Kurzweil offers is $f(x) = 2 \cdot 10^{16}$ (Kurzweil, 2004 [18], p. 393), where 10 is the constant ratio of the function. This means that as the input increases by 1, the output value will be the product of the base and the previous output, regardless of the value of, here, $a = 2$.

One of the assumptions, according to Kurzweil, is that computer power (bps) is a linear function of the knowledge of how to build computers (Kurzweil, 2004 [18], p. 393). If the aim is to build AI after the human brain, then the exponential growth of information technology and hence, the singularity, is directly dependent on neuroscience (and this relationship is non-exponential, not even polynomial > 1). Moreover, Kurzweil poses that this knowledge accumulates and increases and by that he ignores that knowledge can fade, decay, that people forget ancient skills and methods, do not know how to repair legacy software.

We would achieve "one Human Brain capability" by 2023 (Kurzweil, 2004 [18], p. 393). Question is, whose brain? The average brain, the median brain, the least intelligent percentile? What falls under 'capability?' Bytes per second seem hardly relevant for effective problem solving, whether artificial or human. Arguably, do we have that brain-like computer today? That is a matter of perception. If one conceives of the brain as a cellular automaton, then yes, we do. If we think of the brain as an information processing organ that on its wetware integrates memory, affect, skills, thinking, thermodynamics, metabolism, and much more, then we are far off. Thus, the best we can do is to let the most humanlike robot be handled by human intelligence and measure the perception of its users, see if they feel the robot emulates human performance.

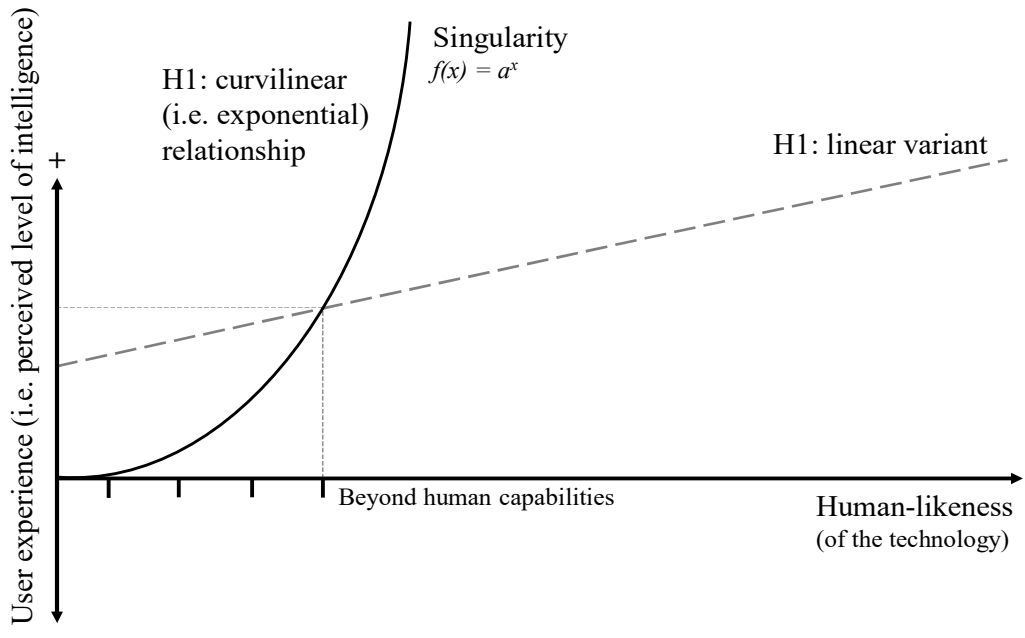

Figure 2: Human-likeness hypothesis, singularity variant: Positive user experience grows with increased human-likeness (exponential conception).

Singularity thinking assumes that with every developmental step on the *x*-axis (Figure 2), the increase in technological capabilities becomes bigger (the *y*-axis). One of the oddities of exponentials is that what determines the shape of the curve is the size of developmental steps, the 'intervals.' With the approximately 6-year interval and the shortening of that period later, singularity thinkers apparently are in doubt of equal or unequal intervals between technological increments.

Nonetheless, in assuming exponential increase, let us regard user experience and human-likeness as a parametric equation where time is the parameter, i.e., $x = x(t)$ and $y = y(t)$, and elimination of $t$ in the parametric equation gives the exponential or other explicit form of function between $x$ and $y$. Note that, for exponential growth, the above implies that either $x$ or $y$ does not follow exponential growth with time (which is probably assumed as linear).

At least we know that the growth curve itself should be super-linear, although that may turn out to be quadratic, which is not what singularity theorists want. In committing themselves to exponentials, singularity thinkers risk that the rate of change of the quantity is not proportional to the quantity itself; that is how rigorous the definition of exponential is and any other rate empirically obtained is not exponential, hence, not indicating the singularity.

Furthermore, because 'the singularity' is near (Kurzweil, 2005 [19]) and thus, not there yet, human performance remains the standard for any sub-singular technology we at present have. Moreover, what the technology's 'capability' is and how big the next jump may be, is human judgment. Thus, the $y$-axis represents human experience of technological advance, which in singularity thinking is limited to the user perception of 'machine intelligence' (whatever the definition may be).

To approach singularity hypothesizing empirically, we have to make specific choices, which are arbitrary because the original ideas are so elusive. Growth is exponential and to start as simply as possible, that means that the value of the function is a constant (although no argument is raised why it would be). If this value is not fixed by some parameter but varies with $x$, then by definition, growth is not exponential. On the basis of Kurzweil's (2004 [18], p. 393) own equation derived from curve fitting of (unavailable) data, what we should find is:

$$f(x) = 2 \cdot 10^{16} \quad (1).$$

However, one could argue that Kurzweil's conception is on a grand scale of neural networks over years achieving humanlike performance in general and is not about a limited number of robots, interacting with a few people for a few minutes. Be aware, however, that exponential increase by definition is scale-independent and should exhibit the same curve although the precise rate may be different. Nonetheless, we could help the theory by assuming a less specific equation, which is exponential still:

$$f(x) = a^x \quad (2)$$

with $a \neq 1$ and $a > 0$, being the constant and base of the function and $x$ being the variable. (Remark: $a^x$ and $e^{bx}$ are the same representation, as $e^{bx} = (e^{\ln a})^x = a^x$ for $b = \ln a$). Different from (1), equation (2) does not say what the value of constant $a$ might be. Mathematically, the base of an exponential function often is found to be the 'transcendental number' called $e$, approximating the value of $e \approx 2.72$. Slightly more precise than (2), equation (3) predicts:

$$f(x) = e^x \quad (3)$$

with $e \approx 2.72$. (Alternatively, assuming a general form of $e^{bx}$ (neglecting scaling and offset) may turn out to represent the singularity better with $b$ determining the speed of evolution).

Equation (2) is our most lenient test prediction for the singularity variant of H1 (Figure 2), emphasizing that more human-likeness of the technology increases user experience (here, mainly its 'level of intelligence') with ever-bigger steps.

**1.3 Uncanny Valley theory**

In line with ME and CASA, Henschel, Laban, and Cross (2021) [10] point out that robots providing social cues may be a strong indication for the user how to interact with the machine, but these authors also warn against the danger to fail the user's expectations and fall into the 'Uncanny Valley.' Uncanny Valley theory is on the downside of adding ever more humanlike characteristics to the machine and so expects a limitation to the increase of positive user experiences. ME and CASA commonly ignore this older theory and focus on its first part alone, assuming that more human-likeness designed into the machine increments positive experiences. Uncanny Valley goes further, however, positing a limit to the increase in appreciation of human-likeness or at least a pothole in the road.

While "humanlikeness" was the most common theme related to researchers' descriptions of robot designs, participants did not focus on how humanlike a robot was, and humanoid robots were not evaluated most favorably by them. (Lee, Šabanović, & Stolterman, 2016) [21]

Mori (1970 [34]; Mori, MacDorman, & Kageki, 2012 [35]) introduced the Uncanny Valley (UV), stating that familiarity, later replaced by 'affinity' of a user with a lifelike robot decreases once the robot comes close to being human but not quite, for instance, when mismatches occur between human appearance and awkward movements (Saygin et al., 2012) [42]. That would excite feelings of eeriness, dislike, violation of expectations, category mismatches, etc.: The number of user-experience variables seems to change per Uncanny-Valley study (Kätsyri, Förger, Mäkäräinen, & Takala, 2015) [13]. According to Mori [34], the just-not-human-enough robot would be reminiscent of corpses and zombies and so provoke aversive effects (Figure 3). That does not always have to be a bad thing, for example, if to ward off intruders (Lee, Šabanović, & Stolterman, 2016) [21]. Note that UV has a bipolar conception of affect, where less affinity is more eerie (affinity = 1 – eeriness). We later shall see that maintaining two parallel unipolar dimensions will fit the UV presumptions better.

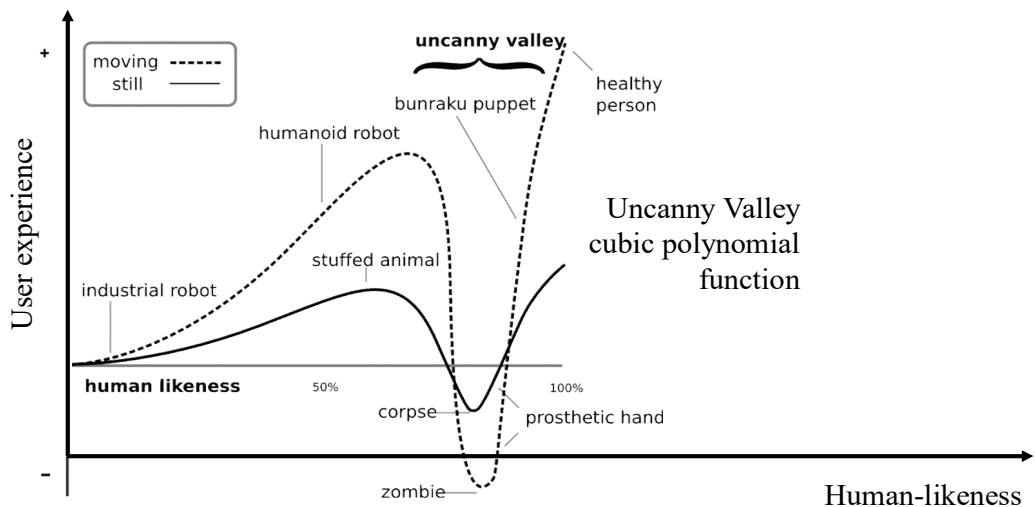

Figure 3: Uncanny Valley expects positive user experience to become negative (bipolar conception) with too much human-likeness.

There are many explanations forwarded to account for uncanny effects (Zhang, Li, Zhang, Du, Qi, & Liu, 2020) [50]; Tinwell, Nabi, & Charlton, 2013 [44]; MacDorman, Green, Ho, Koch, 2009 [29]) but in itself the Uncanny Valley (e.g., Rosenthal-von der Pütten & Krämer, 2014) [41], obviously, goes against Media Equation and CASA, which assume that the user experience just improves with more human-likeness – provided the cue is not deliberately negative such as rudeness or negative feedback.

Similarly, singularity theorists do not worry much about the negative effects of technology's increased capabilities (Kurzweil, 2005, Chap. 2) [19]) and most of their attitude is optimistic: They (implicitly) expect positive user experiences from humanlike and superhuman technologies.

In resonating UV, our H2 advocates a simple *Human-likeness: too much of a good thing.* The power function depicted in Figure 3 already tells this is a non-linear equation. If $x$ were squared ($x^2$), the graph would describe a parabola, which it obviously does not. The polynomial that UV typically forwards is $x$-cubed ($x^3$), describing the rise-dip-and-rise of Uncanny Valley hypothesizing. Thus, for ME and CASA, we predict a linear increase and no power function (~ $x$) or maybe a quadratic relation (~ $x^2$) (a parabola or inverted U) if there is an optimum (which they do not mention). For UV, we predict a dual-extremum (e.g., representable in a power series with at least cubic behavior (~ $x^3$)), while for singularity, we expect an exponential growth curve $a^x$, $e^x$, or $2 \cdot 10^{16}$.

Although Mori (1970) [34] predicted that an entity bearing a very close resemblance to humans would risk eliciting cold, eerie feelings in viewers (MacDorman & Chattopadhyay, 2017) [27] and although this view is accepted by a wide circle of designers and researchers, such responses lack strong support from empirical research so far (MacDorman, 2006 [26]; MacDorman & Ishiguro, 2006 [30]; Tinwell, 2009 [43]). Quite like the followers of ME and CASA, who have not doubted the importance of human-likeness, have followers of UV theory hardly ever doubted that the effect will occur or that it is important when it occurs. It seems, then, that in the design of humanoid robots, avatars, and AI, human-likeness is the gold standard only kept in check by uncanny effects that potentially may happen. In spite of this double trend, the current article is in doubt of both.

We argue that Media Equation is an important theory because many psychologists and designers (of Artificial Intelligence) believe in human-centric design and human beings as the gold standard of performance and experience. However, Media Equation is an unimportant theory because it describes episodes of human experience of media (including robots) that do occur but are rare and short-lived, happening in some people at some moments in time.

We argue that Uncanny Valley is an important theory because many engineers who do not believe in human-likeness emphasize how too lifelike a robot will scare and appall its users, reminding them of corpses, 'zombies,' or psychopaths (Tinwell, Nabi, & Charlton, 2013 [44]). However, Uncanny Valley is an unimportant theory, because empirically, those effects hardly occur, are not long-lasting, happen only for a handful of people in very specific cases (cf. MacDorman & Entezari, 2015) [28].

**1.4 Task contingency**

Human-likeness and eeriness are in the center of attention of robot theorizing. Like any other technical system, however, social robots also are evaluated for the functionality they provide. The quality of the service offered, the performance of the task, proper software engineering, the ease of interaction, all these aspects weigh in on the user experience and may affect whatever level of human-likeness the system has or how uncanny the machinery may be. Quite like any other interactive system (cf. McNamara & Kirakowski, 2006) [31], functionality and usability should be of concern for designing the user experience of a robot. "For a commercial robot product, it is the achieved UX in the natural context when fulfilling its intended purpose that will determine its success" (Lindblom, Alenljung, & Billing, 2020) [25]. We believe that affordances of the robot should be relevant to and aligned with the goals of the task at hand for the user to experience 'affinity' with the machine (Konijn & Hoorn, 2017, proposition 4) [17]. That a robot has wheels is of no concern to having a conversation but an NLP module is.

Figure 4 shows that what the robot affords (e.g., face recognition) should be relevant to the task (e.g., to check someone in), leading to higher intentions to use the machine (Van Vugt, Hoorn, & Konijn, 2009) [45]. Figure 4 also shows that additionally, affordances (action possibilities) may exert direct effects on user engagement (whether feeling involved or at a distance) (ibid.) [33].

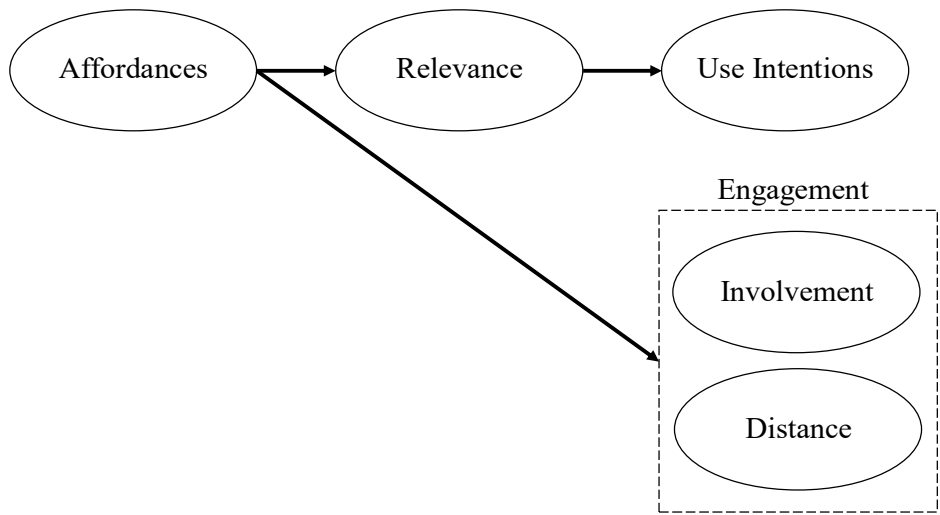

Figure 4: Task-related factors may determine the user experience of a robot.

The difference with the previous theories, ME, CASA, and UV, is that we focus on affordances rather than realistic social cues (whether in appearance or behavioral). For ME and CASA, the increase in realism heightens involvement. For UV, the increase in realism increases distance (i.e. eeriness, uncanniness). Singularity ideas are somewhat closer to ours as 'intelligence' would count as an affordance, increasing involvement (i.e. inspiring enthusiasm). However, none of the previous theories include relevance to goals and concerns nor use intentions in their theorizing.

The four Figures 1-4 illustrate three hypotheses: H1 is *Human-likeness: the more the better*, claiming that an increase in human-likeness improves the user experience. In its most parsimonious version, the ME/CASA variant of H1 predicts a linear slope coming from a significant intercept, with a direct effect of perceived realism on involvement.

H1 in the singularity variant predicts non-linearity, where human-likeness of technology positively impacts user experience in an exponential manner ($a^x$, $e^x$, or $2 \cdot 10^{16}$), which in our conception is the perceived level of affordance (i.e. technological intelligence), exciting user involvement.

H2 is derived from Uncanny Valley: *Human-likeness: too much of a good thing*. There supposedly is an optimum in human-likeness so that user experience drops when human-likeness becomes too high but is not human-identical yet. The predicted curve should be a cubic polynomial ($x^3$) of perceived realism (in appearance, in behavior) reducing the experience of involvement in the comparison between the most humanoid robot and a real human being. Note that UV theory regards a drop in involvement ('familiarity,' 'affinity') as a rise in affective distance (i.e. eeriness, uncanniness).

H3 says that appreciation of *Human-likeness is task-contingent.* Task-related experiences are key for H3, saying that affordances perceived to be (ir)relevant to task execution as well as intentions to use the robot are influential for the user experience even if H1 and H2 hold true. Figure 4 shows that unlike UV, involvement and distance are not considered bipolar but two unipolar scales that may happen in parallel: Appeal and appall can occur simultaneously. The robot's facial expressions may be fascinating, yet repulsive.

Robot research commonly comprises of one or sometimes two robot conditions to compare human-human with human-robot interaction (e.g., De Graaf, Ben Allouch, & Van Dijk, 2015 [3]; Moshkina, Trickett, & Trafton, 2014 [36]; Złotowski, Sumioka, Eyssel, Nishio, Bartneck, & Ishiguro, 2018 [51]). The current study offers a comparison among multiple robot conditions, human-like and machine-like, as well as a human-human condition in an experiment that tested assumptions of human-likeness and task execution.

## 2 METHODS

### 2.1 Participants and Design

After obtaining approval from the institutional Ethical Review Board, in the last months before social distancing took effect, we administered a field experiment at the Hong Kong Productivity Council (HKPC) as well as Hong Kong Science Park, using four different agencies to register participants for three different events (i.e. seminars). Before taking our questionnaire, voluntary participants ($N = 89$; $M_{age} = 42$, $SD_{age} = 13.58$; 54 female) were randomly assigned to one of the conditions in a 4 (Agency: Iwaa robot vs. Sophia robot vs. Sophia avatar vs. Human confederate) × 2 (Eye contact: looking-at, looking-away) between-subjects experiment. Note that the factor level 'looking away' was added for the two robots alone, not for the avatar and not for the confederate. The questionnaire was taken after check-in and before the start of the seminar. Informed consent was obtained formally from all participants. In completing the questionnaire, participants received a lucky-draw opportunity to win a Hanson's *Professor Einstein*™ robot.[1] Details of participant distributions can be found in the Technical Report (see online Supplemental File).

### 2.2 Procedure

To register for a seminar, participants were guided into one of four lanes and were serviced in English by the corresponding agency (Smart Dynamics' Iwaa, Hanson Robotics' Sophia, Hanson's Sophia's Avatar, or a human confederate). Apart from helping to check in, the affordance we manipulated was eye contact (looking at or away from the user) because it may enhance not just the assessment of the quality of task execution (i.e. identifying the seminar attendant) but may enhance the experience of human-likeness as well or when looking away, increase uncanniness. Participants interacted with the agency and after check-in, they filled out an online questionnaire in the Qualtrics environment for administration of surveys and experiments. Participants received a tablet to do so or scanned a QR code to work on their mobile phones. Figure 5 and 6 provide a view of the experimental set-up. Consult Appendix 1 of the Supplemental File for the various layouts at the three different locations.

---

[1] hansonrobotics.com/professor-einstein/

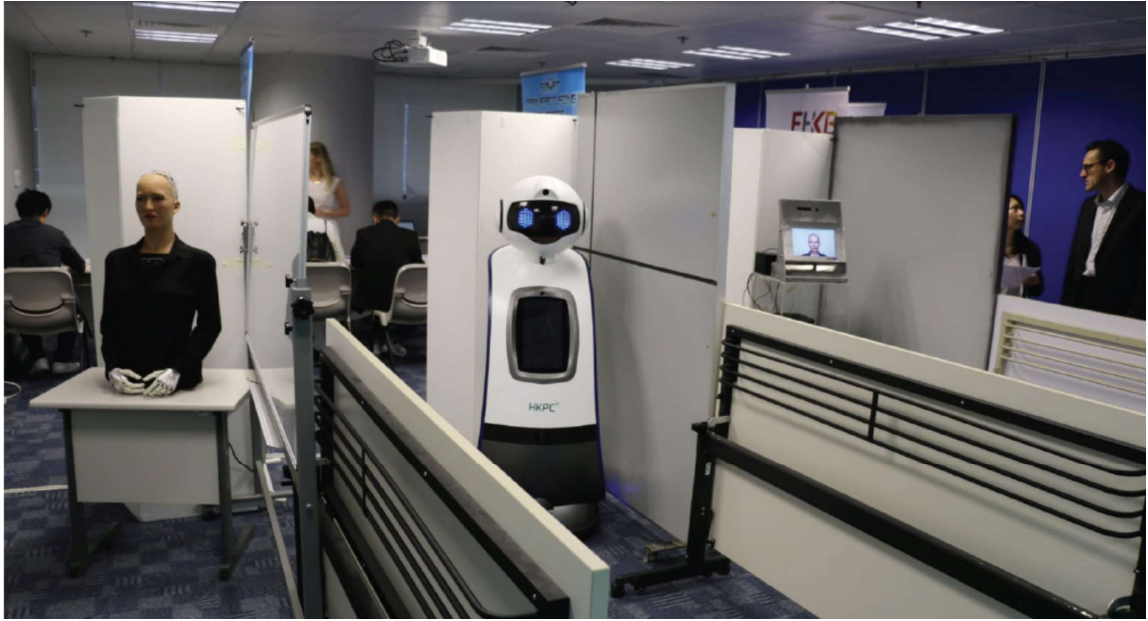

Figure 5: HKPC set-up: Sophia robot, Iwaa robot, Sophia avatar, and female confederate. Participants fill out questionnaires in the background.

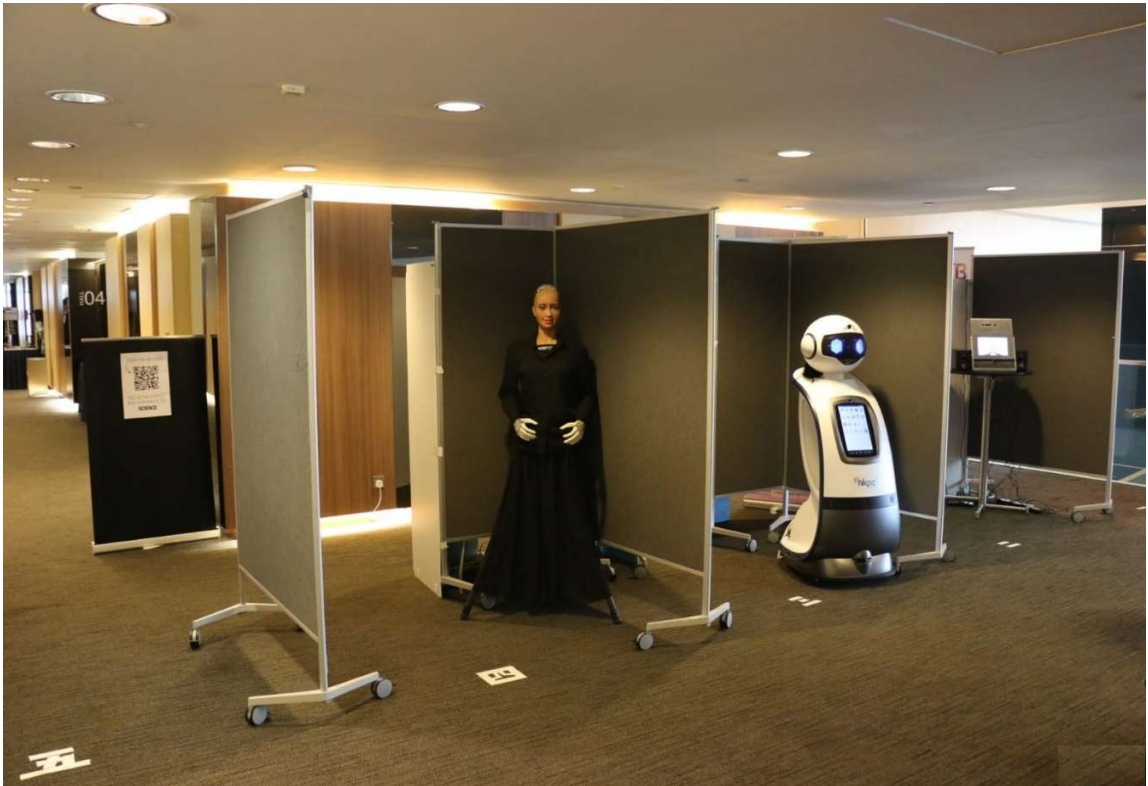

Figure 6: Experimental set-up at the Science Park location. Lane 1 had the human confederate. Participants left at the back. The QR code led to the questionnaire.

As to the experimental layout, there were differences due to the available locations and practicalities at HKPC and Science Park. The robots at HKPC were in a different lane from Science Park. At HKPC, Sophia's upper body stood on a table; in Science Park, it was mounted on a tripod, wrapped in a skirt. At both locations, participants were brought in one by one by the lead experimenter from the right side of the room to avoid them from seeing the robots or other participants interacting. The lead experimenter would walk in between participant and robot lanes, blocking the view on the robots as much as possible but even so, participants might have caught glimpses of robots and people interacting, perhaps being aware of experimental conditions. In the **Analysis and Results**, however, the effects of Event (i.e. the three seminars) were not significant so harm to experimental validity should be little.

The dialog and protocol the agency followed were fixed, including that of the confederate. In one condition, the robot (not the avatar, not the confederate) looked at the participant, in the other, the robot looked away to provoke effects that possibly sustained Media Equation or Uncanny assumptions. For all four agencies, after interaction, participants filled out a 61-item structured questionnaire (Supplemental File, Appendix 2) on how they experienced the agencies, ending on demographic information. Items on the questionnaire were presented in blocks with pseudo-random sequences of items within blocks, different for each participant. The questionnaire was provided in both English and Cantonese. Upon completion, the participant was brought into the seminar room.

## 2.3 Apparatus and Materials

The four agencies were Smart Dynamics' Iwaa robot, Hanson's Sophia robot, Hanson's Sophia on-screen avatar, and a female confederate (24 years old, Cantonese, 1.60m height, higher education) (Figure 7). Smart Dynamics' Iwaa is a human-tall rectangular robot with a round head, carrying a black screen with two blue blinking eyes. It has limited facial expressions and head movement. Hanson Robotics' Sophia is a human-size and highly humanoid robot with a realistic face delivering various facial expressions, moving its head and arms. As a research platform, it can be integrated and customized through the Hanson API.

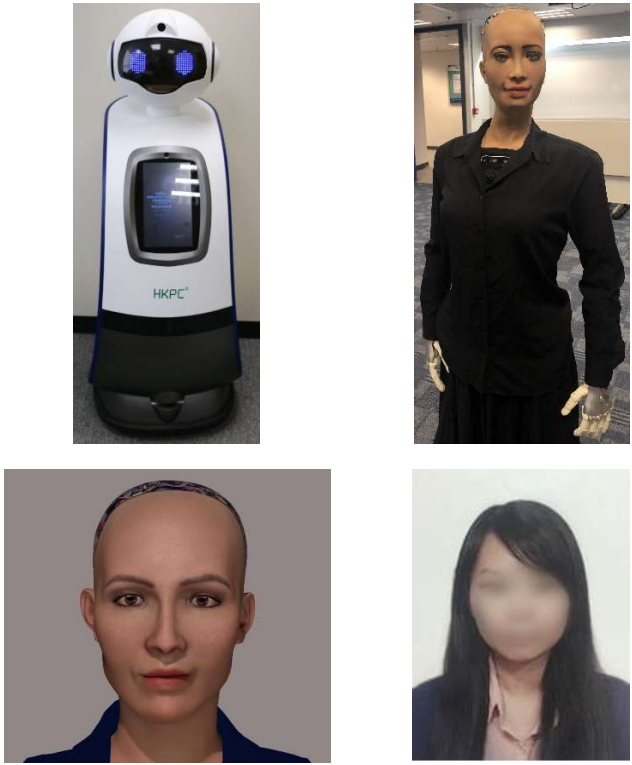

Figure 7: Iwaa robot, Sophia robot, Sophia avatar (on-screen), and human confederate (blurred).

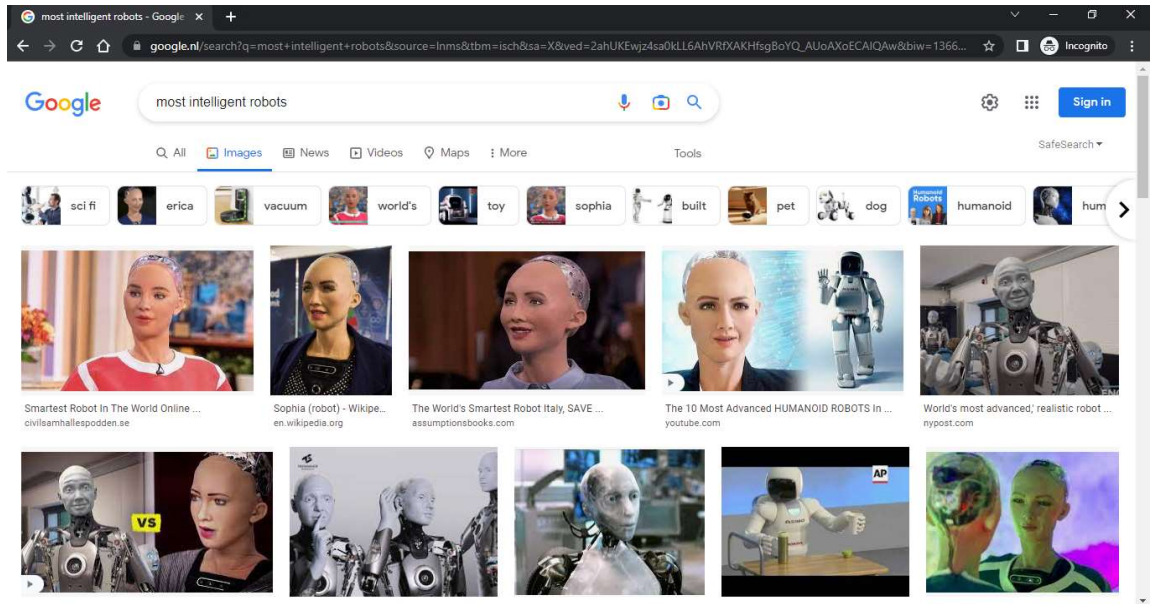

Figure 8: Sophia is perceived as one of the most advanced robots, Iwaa is not.

When Googled for "most intelligent robots," Sophia hits the top ten of all ranking lists whereas Iwaa obtains zero hits (Figure 8). Also in the PR and marketing of the machine, Sophia is advertised as 'personifying our dreams for the future of AI' and "one of the key members of the founding SingularityNET team."[2] Whether wild exaggeration or not, publicly, Sophia is perceived as having human capabilities and thus, should be capable of provoking singularity-type responses. Kiesler, Powers, Fussell, and Torrey (2008) [14] found differences between physical and non-physical agents (i.e. robots vs avatars) so that we inserted a condition of Sophia as an avatar as well.

In spite of the huge claims made by researchers and developers, no stota AI is capable of autonomously managing a real-life check-in so that the three artificial agencies were handled in remote control. The robot operators were invisible to the participants who had no reason to believe the agencies were not autonomous. The human confederate was trained to hold strict protocol like the robot operators did. The check-in protocol of the agencies consisted of greeting, asking the English name, asking the company name, asking whether the participant needed extra help, and closing with saying goodbye and to proceed (for interaction flowcharts and dialogs, see Supplemental File).

## 2.4 Measures

To test hypotheses of anthropomorphism, uncanniness, and task-related experiences, a structured questionnaire queried the following dimensions: Perceived human-likeness (e.g., has feelings), Involvement (i.e. experiencing warmth, engagement), Distance (i.e. eerie, uncanny), Use intentions (e.g., wanting to work with the agency), Task relevance (useful, worthwhile), Task quality (job well done, good help), and Parallel-processing (of cognition and affect). Perceived human-likeness was flanked by three sub scales: Anthropomorphism (e.g., has personality, a soul), Theory-of-mind (e.g., recognizes how I feel), and Perceived realism (e.g., looks human); this way, human-likeness was not queried by a mere '… looks like a human' alone. The cognitive aspect of Parallel-processing was called 'High-path' (e.g., 'The interaction made me reflect on what was going on') and affective processing 'Low-path' (e.g., 'I responded intuitively…'), indicating the neurological pathways supposedly being in operation (Konijn & Hoorn, 2017) [17]. In total, the questionnaire consisted of 13 theory-related scales (Supplemental File, Appendix 2).

---

[2] https://www.hansonrobotics.com/sophia/; https://blog.singularitynet.io/the-dao-of-sophia-14034855929e

To control for possible confounds of the robot being new and beautiful, affecting the experience, an Aesthetics (e.g., looks nice, pretty) and a Novelty (e.g., new, original) scale were devised. Demographics included Gender, Age, Education, and earlier experiences with robots. For exact descriptions of each variable, see the Supplemental File).

Apart from demographics, items were Likert-type statements, including counter-indications, followed by a 6-point rating scale (1 = strongly disagree, 6 = strongly agree). Items within blocks were pseudo-randomly presented to each participant.

Before reliability analysis, we reverse-coded the counter-indicative items. For each scale, we calculated Cronbach's α to determine convergent validity. After items deleted, where there were two items left, we calculated Spearman-Brown correlations. After reliability analysis, we did Principal Component Analysis (PCA) to determine divergent validity of our measurements. An elaborate account of the reliability analysis can be found in the Supplemental File.

In brief, most variables of theoretical interest obtained $.85 \leq$ Cronbach's $\alpha \geq .95$, indicating good to very good internal reliability. Perceived human-likeness did not perform too well ($\alpha = .57$) and after deletion of two items, Spearman-Brown indicated that $r = .63$, which is just acceptable. The items of the (shortened) scales went up for PCA (Promax rotation) to find 13 components in a free-fitting format. Out of 13 variables, the PCA extracted 11 significant factors that explained 80.45% of the total variance.

With 11 components, two scales were scattered all over the other variables, so we decided to remove the theoretically less important factors such as Aesthetics, Novelty, and the poorly performing factor of High-path (with a wide spread and a mediocre internal consistency of $\alpha = .62$). The PCA showed that participants saw Use intentions, Task relevance, and Task quality as aspects of one and the same dimension, so we merged the respective items into one container concept named Task-related experiences. We also removed the weaker Perceived human-likeness items ($r = .63$) and based on the PCA results, aggregated the related scales (Anthropomorphism, Theory-of-mind, and Perceived realism) into one measure of Perceived human-likeness (revised).

In sum, we started with 13 scales and PCA extracted 11 components. Then, for various reasons, we removed Aesthetics, Novelty, Perceived human-likeness, and High-path. We took Anthropomorphism + Theory-of-mind + Perceived realism together and did the same for Use intentions + Task relevance + Task quality. This drastic simplification of our measurement dimensions left us with 5 scales to perform a second PCA on (free fit).

Distance and Involvement separated nicely, and unlike before, Involvement formed its own cluster. With regard to Perceived human-likeness (revised), two more of its items loaded onto this component as compared to the first run. Also, both items "Agency is alive" and reverse-coded "Agency is dead material," "Agency has free will," "has a soul," and has "self-control" clustered into Perceived human-likeness (revised). Therefore, we updated Perceived human-likeness (revised) with these items in.

Except for those that were reverse-coded, all items of Use intentions, Task relevance, and Task quality fell into one component, which we termed Task-related experiences, excluding the counter-indications, which fell outside this component. The counter-indications of Task-related experiences formed a separate component, which we discarded.

Low-path items clustered neatly together, whereas in the first PCA already, High-path items spread all over components. It may mean that affective aspects of the interaction (e.g., 'uncanny') were clearly present for the participants, whereas reflections about that experience were less clear. Maybe the High-path items were a little more complex, inducing confusion (e.g., "During the interaction, I wondered if the conversation was scripted").

In view of Konijn and Hoorn (2017) [17], Low-path results are not too interesting if they are not accompanied by a High-path dimension. Therefore, we decided to once again simplify our model and work with the measurements that had *and* good reliability *and* were theoretically intact.

In all, the four remaining (shortened) measurement scales achieved very good reliability in the final Cronbach analysis: Perceived human-likeness (revised) (Cronbach's α = .95), Involvement (Cronbach's α = .93), Distance (Cronbach's α = .95), and Task-related experiences (Cronbach's α = .95). Likewise for the controls Aesthetics (Cronbach's α = .87) and Novelty (Cronbach's α = .81). Details are provided in the Supplemental File.

We then calculated the average scores for each scale and performed outlier analysis, clustering on the factors Agency and Eye contact. There was a total of 16 outliers, spreading over different variables. After checking their answers, we could not conclude for an acquiescence response bias, except for participant 64 and 86, which were removed from our data set. We then created two data sets – one with outliers included ($N = 87$) and the other with outliers excluded ($n = 79$). We performed our effects analysis on both these data sets.

## 3 ANALYSIS AND RESULTS

In all of the analyses reported next, none of the background variables (e.g., Age, Gender, Event) exerted significant effects (Supplemental File). That Event rendered no significant effects dismisses the suspicion that participants experienced the set-up of robots systematically different at different seminars. Perhaps participants saw the other robots, perhaps they noticed that Sophia was on a table and at another location on a tripod covered with a skirt but that did not affect their experiences in a consistent manner.

We ran a 4 Agency × 2 Eye contact General Linear Multivariate Analysis (MANOVA, Pillai's Trace) on the mean scores for Perceived human-likeness, Distance, Involvement, and Task-related experiences with Aesthetics and Novelty as covariates. We did this for $N = 87$ (with outliers) and $n = 79$ (without) separately. Neither effects on Involvement or on Task-related experiences nor any interaction with any of the dependents was significant.

Yet, for $N = 87$, we did find significant multivariate effects of Agency ($V = .80$, $F_{(3,77)} = 5.04$, $p = .000$, $\eta_p^2 = .27$) and of Eye contact ($V = .36$, $F_{(2,77)} = 7.50$, $p = .000$, $\eta_p^2 = .36$). Univariate analysis showed that Agency and Eye contact both exerted a significant main effect on Perceived human-likeness (Agency: $F_{(3,77)} = 26.07$, $p < .000$, $\eta_p^2 = .58$; Eye contact: $F_{(1,77)} = 13.85$, $p < .000$, $\eta_p^2 = .20$). Therefore, we ran 6 independent samples *t*-tests for the effects of Agency on Perceived human-likeness and after Bonferroni correction (α = .05/6 ≈ .0083), we found that only the differences with the Human confederate were significant, favoring the Human confederate: Sophia robot vs Human confederate ($t_{(46)} = -8.12$, $p = .000$), Sophia avatar vs Human confederate ($t_{(26)} = -6.46$, $p = .000$), and Iwaa robot vs Human confederate ($t_{(39)} = -7.00$, $p = .000$). Figure 9 shows means and standard deviations. For the effect of Eye contact on Perceived human-likeness (looking-at: $M = 2.98$, $SD = 1.49$; looking-away: $M = 3.42$, $SD = .40$), independent samples *t*-tests rendered non-significant results: $t_{(79)} = -1.20$, $p = .23$. For $n = 79$, we found similar patterns of results (Supplemental File). By no means was even 'the most advanced robot' nearing the human being, who 'unnaturally' followed robot protocol.

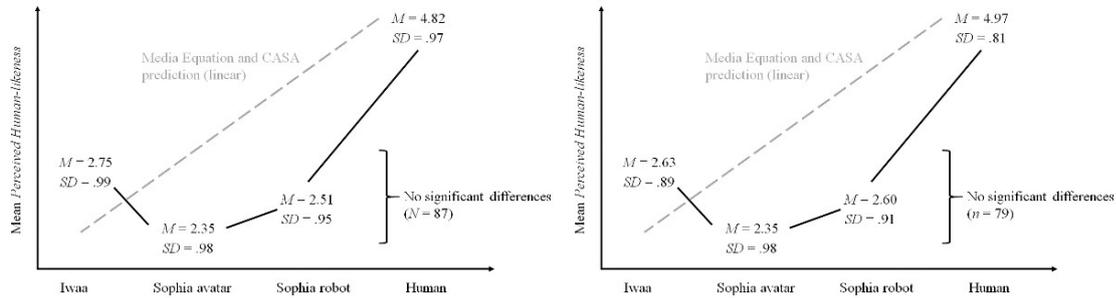

Figure 9: Mean Perceived human-likeness differentiated by Agency (left: *N* = 87, right: *n* = 79).

The 4 Agency × 2 Eye contact GLM Multivariate (Pillai's Trace) also revealed effects on mean Distance with *n* = 79. As mentioned earlier, univariate analysis showed a significant main effect for Distance ($F_{(1,69)}$ = 3.10, *p* < .035, $\eta_p^2$ = .16) and accordingly, we ran 6 independent samples *t*-tests with α = .083, according to Bonferroni. Figure 10 shows the averages. Again, significant effects happened only in comparison with the Human confederate. The robots raised significantly more feelings of Distance than interacting with a human being did: Sophia robot vs Human confederate ($t_{(41)}$ = 5.42, *p* = .000), Sophia avatar vs Human confederate ($t_{(25)}$ = 3.30, *p* = .000), and Iwaa robot vs Human confederate ($t_{(34)}$ = 2.80, *p* = .000). Informal analysis of the averages showed a small peak in Distance for the most humanlike Sophia (Figure 9), which would support UV ideas but as said, all differences were insignificant.

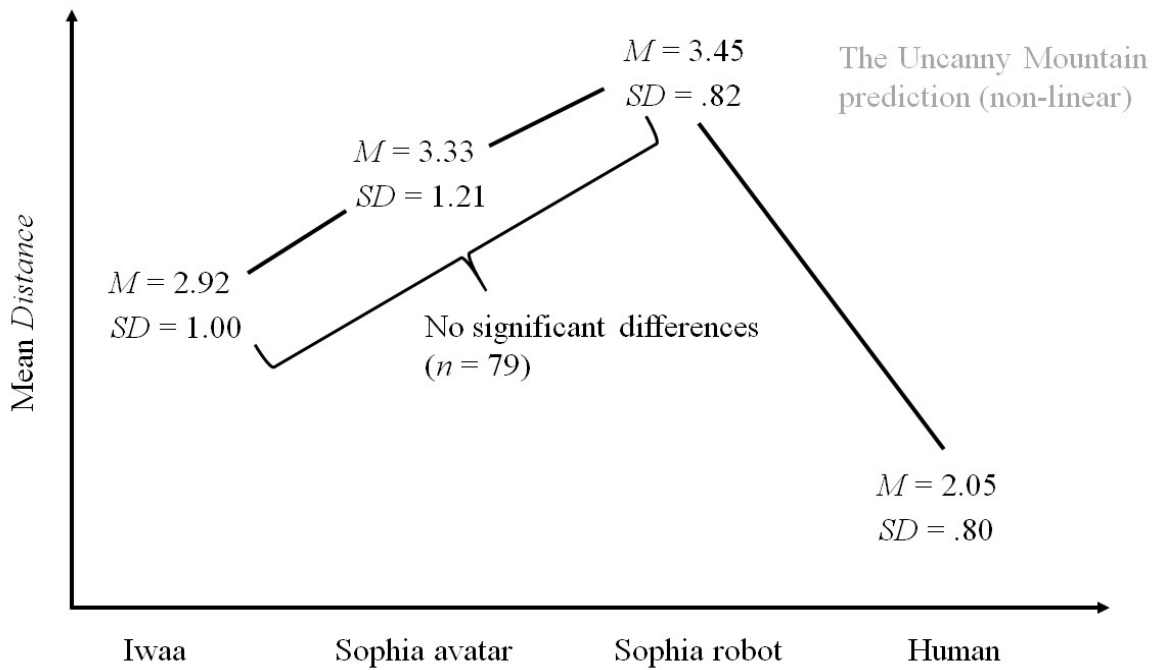

Figure 10: Mean Distance differentiated by Agency (*n* = 79).

*Control variables.* For $N$ = 87, multivariate effects occurred for mean Aesthetics ($V$ = .28, $F_{(1,77)}$ = 5.16, $p$ = .001), which covaried with Distance ($F_{(1,77)}$ = 14.85, $p$ < .000, $\eta_p^2$ = .21), Involvement ($F_{(1,77)}$ = 11.86, $p$ < .001, $\eta_p^2$ = .18), and Task-related experiences ($F_{(1,77)}$ = 13.40, $p$ < .001, $\eta_p^2$ = .19). Mean Novelty ($V$ = .27, $F_{(1,77)}$ = 5.00, $p$ = .002) covaried with Involvement ($F_{(1,77)}$ = 13.88, $p$ < .001, $\eta_p^2$ = .20) and Task-related experiences ($F_{(1,77)}$ = 7.47, $p$ < .001, $\eta_p^2$ = .18). For $n$ = 79, these patterns of results were the same. Experiences of Aesthetics mitigated feelings of Distance ($r$ = -.69***) and both Aesthetics and Novelty enhanced experiences of Involvement ($r$ = .65*** and $r$ = .61***, respectively) as well as Task-related experiences ($r$ = .66*** and $r$ = .55***, respectively).

To tease out ME and CASA inspired behaviors in the user, we devised the Eye-contact condition for Iwaa and Sophia robot, which should lead to more involvement and less feelings of distance, in line with the idea (H1) that more human characteristics increase engagement. Indeed, some studies claim that eye contact builds trust between robots and humans (Kiilavuori et al., 2021 [15]; Kompatsiari et al., 2019 [16]).

To test the assumption, we performed 4 separate linear-regression analyses (Method Enter) of Eye contact on Involvement, on Distance, and on Task-related experiences. The association between Eye contact and mean Involvement ($R^2_{adj}$ = -.006, $\beta_{sc}$ = -.086, $t$ = -.74, $p$ = .465) was not statistically significant, nor was that between Eye contact and Distance ($R^2_{adj}$ = -.013, $\beta_{sc}$ = .021, $t$ = .18, $p$ = .857). However, it did explain changes in Task-related experiences ($R^2_{adj}$ = .045, $\beta_{sc}$ = -.243, $t$ = -2.07, $p$ = .043). With the agency making eye contact, task-related experiences such as task relevance, task quality, and use intentions improved significantly by 2.43 scale points. The results for Involvement and Distance are unexpected by ME and CASA; the effects on Task-related experiences could be interpreted as support of ME and CASA although none of the theorists ever mentioned task-contingency of their theories.

We also performed a linear regression analysis to check whether the change of mean Perceived human-likeness had a beneficial impact on Involvement and Distance. Results showed that Perceived human-likeness significantly explained the increase of Involvement ($R^2_{adj}$ = .066, $\beta_{sc}$ = .281, $t$ = 2.47, $p$ = .016) and the decrease of Distance ($R^2_{adj}$ = .24, $\beta_{sc}$ = -.499, $t$ = -4.85, $p$ = .000). This finding is in line with ME and CASA expectations.

In trying to accommodate the singularity variant of H1, we ranked the averages for Human-likeness of the four agencies from low to high (Figure 9): Sophia avatar, Sophia robot, Iwaa, Human being. This in itself is a peculiar series as the Iwaa, which never hits the top 10 rankings, outdid Sophia, which is publicly regarded as the closest thing to human sentience and intelligence (Figure 8). If we look at the respective averages 2.35 – 2.51 – 2.75 – 4.82 ($N$ = 87) and 2.35 – 2.60 – 2.63 – 4.97 ($n$ = 79), we see a difference. Outliers excluded ($n$ = 79), the differences between averages are .25, .03, and 2.34. The dent in the increase between Sophia robot and Iwaa (.03) directly defies any assumption of an exponential function. With outliers included ($N$ = 87), however, something special occurs. This time, the differences between averages are 16, 24, and 207, respectively. If we are kind to singularity assumptions and consider 207 being close enough to 208, then these are increasingly larger multiplications of 8: 2×, 3×, and 26×, respectively. May that, perhaps, indicate a glimpse of experiencing the singularity in a real-life situation?

For testing H2, we performed a curve fit with a quadratic and a cubic model to find the polynomial regression curve depicted in Figure 11. Although the selected model fitted our data ($R^2$ = 0.24, $R^2_{adj}$ = .21, $p$ = 000), which may somewhat resemble UV assumptions, not all the coefficients were significant (Supplemental File), which runs against UV ideas.

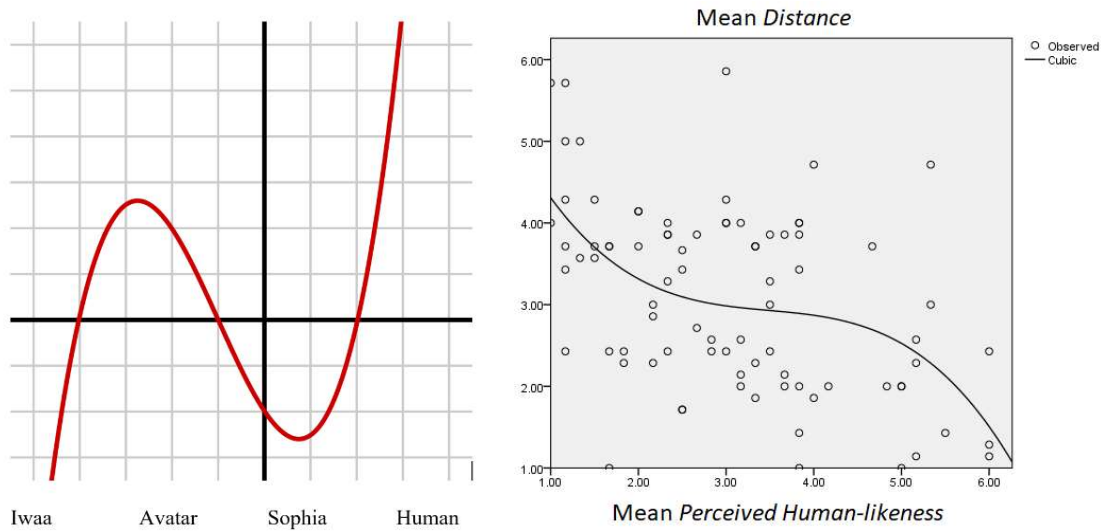

Figure 11: Regression prediction of Uncanny Valley assumptions and non-linear regression of Perceived human-likeness on Distance.

**3.1 Exploratory analysis**

To explore how the artificial agencies performed among one another, off the record, we excluded the Human condition and ran a 3 Agency (Iwaa, Sophia robot, Sophia avatar) × 2 Eye contact GLM on the averages for Perceived human-likeness, Distance, Involvement, and Task-related experiences without Aesthetics and Novelty as covariates. The details of this MANOVA are in the online Supplementary File.

Outliers included ($n = 61$), Eye contact rendered significant effects (Wilk's $\Lambda = .52$, $F_{(4, 57)} = 12.13$, $p = .000$, $\eta_p^2 = .48$). Detailed analysis revealed that the simple effect of Agency on Distance ($F_{(2,56)} = 6.89$, $p = .002$, $\eta_p^2 = .20$) was significant as well as the simple effect of Eye contact on Human-likeness ($F_{(1,56)} = 15.13$, $p < .000$, $\eta_p^2 = .21$) and on Task experiences ($F_{(1,56)} = 4.97$, $p = .030$, $\eta_p^2 = .082$). Three *t*-tests on the Distance rating for the Agencies indicated that Sophia robot ($n = 31$, $M = 3.67$, SD = 1.04) was experienced as significantly more uncanny than Iwaa ($n = 24$, $M = 2.78$, SD = 1.03) ($t_{(53)} = 3.18$, $p = .002$).

Outliers excluded ($n = 56$), the main effect of Eye contact on the combined dependents also was significant (Wilk's $\Lambda = .53$, $F_{(4, 52)} = 12.13$, $p = .000$, $\eta_p^2 = .48$). However, the simple effect of Eye contact on Task-related experiences vanished. Yet, the simple effect of Agency on Distance was significant ($F_{(2,54)} = 4.12$, $p = .022$, $\eta_p^2 = .14$) and so was Eye contact on Human-likeness ($F_{(2,54)} = 10.68$, $p = .000$, $\eta_p^2 = .26$). No significant differences were found when we did pair-wise comparisons between Iwaa, Sophia robot, and Sophia avatar. Although the *p*-value for the contrast between Sophia and Iwaa on Distance was below .05 ($t_{(48)} = 2.20$, $p = .033$), this value failed alpha adjusted for Bonferroni ($p < .01666$).

With the Human condition omitted, participants experienced higher levels of Distance for Sophia robot than for Iwaa. However, the statistical difference was negligible without the outliers. Participants also rated the agent with eyes open as more task-satisfying (in $n = 56$) but as less humanlike (in $n = 61$ and $n = 56$). Take this finding with care, however, as the effect of Eye contact on Perceived human-likeness may be vulnerable to a Type III error (11.06 in $n = 61$ and 10.68 in $n = 56$) (Supplementary File).

## 4 DISCUSSION

Our first variant of H1 (*Human-likeness: the more the better*) was related to Media Equation and CASA: People treat robots as they treat real human beings, while CASA says that they do so even mindlessly (Low path). The more humanlike robots are, the more people treat robots as humans (cf. Moshkina, Trickett, & Trafton, 2014) [36] until robots are developed so well that they equal or even emulate humans (the singularity).

Figure 9 indicates that ME and CASA would expect human-likeness to increase the more human characteristics are encountered. However, the least humanlike Iwaa robot was not perceived significantly different from the more humanlike Sophia avatar or the most humanlike Sophia robot. On average, Perceived human-likeness was even a little better for Iwaa (!). Moreover, looking the participant into the eye also did not add to the perception of human-likeness, not confirming H1. The only support found for ME and CASA assumptions was that Perceived human-likeness significantly explained an increase in Involvement and a decrease in Distance but be aware that the low point of the regression line was with the three artificial agencies, which did not differ significantly among themselves, and the very much higher point of the regression line was with the human confederate, together establishing a significant relationship: More human-likeness was mainly experienced in… the human.

One may wonder why significant effects of eye contact remained absent for the feeling of human-likeness. The social rules applied seem obvious as the protocol for check-in (see Supplementary File) include greeting, asking name, identifying participant, confirming reservation, finding seat, and wishing a good time, which is a complete social script for a check-in. By that, we forced participants in an ME/CASA mode, participants expecting that agencies would respond in a humanlike manner. This nevertheless did not lead to significant effects among the artificial agencies. Only if ME/CASA take back that more and stronger social cues lead to higher perceptions of human-likeness, will the theories stand firm, albeit on very limited ground: Then it is just that humans apply social rules to media, full stop.

The second variant of H1 (*Human-likeness: the more the better*) probed the singularity hypothesis. To make a fair test, our human confederate kept strict robot protocol and so underperformed in human adaptability and exception handling. She yet outperformed all artificial agencies. It seems, then, that two decades after its popularization, robot designers and AI programmers are still far from establishing 'the Singularity.'

Even so, in trying to find confirmation and show goodwill to the theory, we found a regularity in the increase between different stages of technological advance. By including the extreme scores ($N = 87$), differences between averages showed increasingly larger steps of 8: 2×, 3×, and 26×. Maybe we found some indication of singularity experiences.

Alas, the issue is this: If we have at least three measurements (as in our case), then mathematically, it is possible to always find exponential fit (irrespective of goodness of fit). In our case, we can fit the steps between the averages in Human-likeness fairly to an exponential function $6.41694 + 0.165545\ e^{2.36814\,x}$. Quite unbelievably, we would have found a singularity prediction by including a subset of special people (i.e. the outliers) while reversing technological advance (Sophia being less humanlike than Iwaa) as part of our *pro forma* courtesy to the theory.

However, three points that show super-linear increase may not necessarily indicate exponential growth. Given the strict definition of the rate of change, exponential growth is a rather special super-linear function that should be addressed with caution. It is not all too clear that Sophia avatar – Sophia-robot – Iwaa – Human being lie at equal intervals on the *x*-axis. If those intervals are at ordinal scale, which they probably are, then we have found a strictly increasing function *in general*, not necessarily exponential. With unequal intervals between technological advances, the same differences between averages (measures, observations) could be plotted easily as a linear function, supporting ME and CASA, or deliberately antagonistically, as a sublinear function or 'quasi-seminorm,' resisting all variants of H1. Singularity is either untestable or far from near.

H2 (*Human likeness: too much of a good thing*) focused on Uncanny Valley theory, stating that people are scared of robots that are too humanlike. Positive feelings of adding humanlike qualities would drop at near-humanness.

UV expects a cubic polynomial with rising positive experiences when going from a more industrial robot to a more life-like robot and then dropping steeply when the robot comes very or too close to a real human being (in appearance and/or behavior). Reversely, as Figure 11 depicts, emotional distance would be high for an industrial robot, less so for a more humanlike machine, but highest for a robot being (too) close to a real human being. In a for UV theory more advantageous unipolar conception, we could call this finding the 'Uncanny Mountain.' As illustrated by Figure 10, with more humanoid features, emotional Distance (eerie, uncanny) of the participants grew as compared to interaction with the human confederate. Of the three artificial agencies, Sophia robot was deemed most uncanny but not significantly more than the other two, which counters UV hypothesizing.

H3 (*Human-likeness is task contingent*) expected that human-likeness and uncanniness are appreciated or not according to the interactive task the user is engaged in. Therefore, task-related experiences should impact Involvement and Distance but none of the interactions in the GLM Multivariate results were significant. The only significant result we found was that looking at the user explained the experiences related to task execution.

Off the record, we also performed a series of analyses while excluding the Human condition (Section 3.1), just to see how the artificial agencies would do among themselves. Note that these results should be taken with the most caution as we ignored a complete source of variance this way. Irrespective of Agency, the effect of making eye contact on Perceived human-likeness significantly worsened with eyes open (with and without outliers), contradicting H1. For both data sets with and without outliers, Sophia robot was significantly scarier than Iwaa, underscoring H2. If any, open eyes improved Task-related experiences, supporting H3, but only for data with outliers included; an effect of the extreme cases. Thus, looking at the user was experienced as uncanny, particularly for a life-like robot like Sophia but was seen as beneficial for task execution (i.e. for checking in) by some. Taken together, these results would to some extent indicate task-contingency of human-likeness.

If in the official analyses, H1-H3 were disconfirmed then what *did* happen during the interaction? ME and CASA would expect that perceived human-likeness explains the variation in feeling engaged with the agency (see Moshkina, Trickett, & Trafton, 2014) [36]. Indeed, regression analysis showed that Perceived human-likeness significantly explained the increase in Involvement and decrease in Distance. However, Figure 9 and Figure 10 show that the qualities that established those perceptions of being humanlike were significantly more present in the human confederate; much less so among the three artificial agencies, which did not differ significantly among one another on these aspects. Thus, although we found that higher perceived human-likeness related to reduced feelings of distance and enhanced involvement, which is what ME and CASA would expect, in the GLM Multivariate, none of the interactions among dependents were significant so the 'effect' was minor if not negligible. According to *t*-test, making eye contact also did not help to perceive more human-likeness nor had it effect on feeling involved or at a distance. When we regressed Eye contact on mean Involvement, Distance, and Task-related experiences, looking at the user significantly improved experiences of task relevance, task quality, and use intentions. This happened irrespective of whether Iwaa or Sophia executed the task (see the GLM results), thus countering Media Equation and CASA throughout and providing some support for task-contingency.

PCA showed that cognitive reflections were spread over many aspects of the interaction (High-path items clustered with other components) but that emotional experiences were more distinct and clear as related to a number of affective states (Low-path was active), specifically Distance (eerie, uncanny) but also Involvement (warmth, togetherness). If we look into the items that were most prominent for Perceived human-likeness, then it was a compound of perceptions relating to the agency having feelings, personality, 'a soul,' self-control, understanding, a mind of its own, being alive (not dead), and having free will. Therefore, if we want people to perceive an agency as humanlike, it should simulate emotions ('has feelings'), should suggest it has 'a soul,' have moral understanding (note, moral reasoning would be feasible, understanding a situation not), suggest to be 'alive' as in 'not dead,' and show instances of 'free will' (which may be interpreted as showing autonomous behaviors). This would be the requirements list for those who wish to build humanoids and androids that humans consider 'alike.' In other words, the perception of human-likeness has little to do with outer appearance and humanlike realistic looks, which is what Sophia robot capitalizes on.

Yet, while perceiving human-likeness, people also felt 'eerie' and 'scary' as related to the Distance scale. It seems these participants had the feeling they were interacting with the 'living dead,' an experience expected by Uncanny Valley theorizing but not particular for any of the robots as our GLM Multivariate analysis showed.

Uncanny Valley theory says that positive user experiences rise and fall with the increase of perceived human-likeness (appearance, behavior); it is an optimum theory, assuming a cubed polynomial function (Figure 11), the 'valley' being the low point, which we operationalized as little involvement and much emotional distance. However, curve fitting with a quadratic and a cubic solution showed some indication that the selected model fitted our data but not too convincingly (certain coefficients were not significant).

With respect to Aesthetics and Novelty, the experience was better when the agency was seen as new and beautiful. In other words, if old and commonplace, the user experience will wear down.

**4.1 Limitations**

Increasing the ecological validity of a study commonly comes at the cost of experimental soundness as societal and industry partners have their own business processes that cannot be disrupted (here, doing a seminar). That limits what one can do research-wise. Yet, we took caution that people would not see any robot/confederate before entering the experiment room. Participants lined-up outside and then were taken by the lead experimenter to the lane with their check-in agency. The experimenter would walk in between the lane entrances and the participant. If participants noticed some of the other conditions, then that has been a glimpse, not more. If possible, however, researchers may want to organize the experimental layout differently from ours.

Doubts could be casted upon our results, regarding a potential confound in age and gender of our human confederate. At least studies on Uncanny Valley [29] [44] found different responses to the gender of robots/avatars. There are a few things to keep in mind here. First off, not including a human confederate renders ME and CASA invulnerable as there is no comparison to a real being (so how much social script was applied if at all?), while ignoring the Wykowska (2020 [48]) requirement. To reverse the argument, any study into ME and CASA is confounded without any comparison to a human confederate. Second, including a confederate induces right away the infeasibility of a pure replication: It would require a robot portrait of the confederate, which is time-consuming and cost-intensive. Third, the comparison with Iwaa or any other existing industrial robot would go astray. We should have yet another, more mechanistic this time, robot portrait of the confederate. Fourth and most importantly, the robustness of the effects (UV or other) would be at stake. If age, ethnicity, gender, etc. would all make a difference, the underlying theory is in dire straits. The more qualifications an effect needs and the more factors an effect is dependent upon for it to occur at all, the more seldom and the less robust it is, and so the weaker the generalizability becomes of the theories behind. If merely higher-order interactions are significant without the support of main effects, this would underscore how exceptional the effects indeed are.

Unavoidably, the fewer cues to humanness are included, the more gender-neutral a robot becomes (i.e. our Iwaa). Sophia is framed and marketed by Hanson Robotics as a female and people never fail at identifying the Sophia machine as such. Therefore, the gender of the confederate was chosen in line with Sophia robot and Sophia avatar. In that sense, our study excluded the potential confound as much as possible. Future research may look into gendered differences and if confirmed, that would underline our fourth point.

As for test theory, proponents of ME and CASA may note that insignificant findings differ from equivalence, and should be discussed in such terms. Additionally, findings of insignificant differences would not refute theories, but rather fail to support them. However, this would raise a strawman. Nowhere in the paper do we claim statistical equivalence. Contrariwise, we claim media inequality. In fact, our study was directed at finding differences and so rightfully stayed with Fisher and Neyman-Pearson test statistics. It would be ME and CASA employing Bayesian test statistics if they are to claim media 'equation' but to our knowledge, these studies never did. Put differently, ME and CASA often conclude for equality by not finding statistical significant differences, which is unwarranted.

One could argue that theories cannot be refuted but that evidence merely fails to support them. We beg to disagree. It means that ME and CASA escape validation. A remarkable feat of ME and CASA is that these theories inadvertently made the H0 theoretically interesting, a unique aspect never forwarded by the theorists themselves. In human-robot comparisons, conventional (frequentist) statistical techniques (e.g., *t*-test, ANOVA) always compare the data to a null hypothesis (H0). This is what we did. Frequentists can only reject the null or fail to do so. Rejecting the null actually means that in the data significant differences were found between two groups, confirming the alternative $H_a$. However, we did *not* fail to reject the H0 and so correctly concluded for the $H_a$: Humans are perceived as significantly different. Hence, Media Equation and CASA, predicting the H0, were not supported by our data (in fact, they were refuted).

## 5 CONCLUSIONS

The main conclusions we draw from our findings are:

1. Perception of human-likeness has little to do with outer appearance but with internal qualities of the robot, whether programmed or attributed;
2. Media Equation and CASA were not supported by our data. Sophia, regarded as the most human-looking robot ever, was not perceived as significantly more humanlike than Iwaa, not more involving, and was not seen as doing her task better. It was even so that Iwaa had a slightly higher mean score on human-likeness, a reverse effect. The human agency was experienced as significantly different from all robots, Sophia included. This was not a gradual but an abrupt difference;
3. Uncanny Valley theory was not supported by our data. Even for people high or low on feelings of eeriness, the artificial agencies did not significantly differ on experiences of emotional distance or uncanniness. Only informally, when plotting means and SDs in proper order do we see a slight uncanny effect for Sophia robot as expected by Uncanny Valley theory but statistically, that effect is only significant if the Human condition is (illegitimately) excluded from analysis;
4. With regard to task-related experiences, with eye contact, task execution was seen as more relevant and of better quality, making the user more willing to interact. This is independent of the robot. Also in general, the main effect of Agency on Task-related experiences was not significant. The upshot is that the human confederate did not execute her task any better or worse than the artificial agencies. Apparently, people focused on function irrespective of who delivers the goods, whether humanlike or not, whether uncanny or not, and this functionality had little to do with feeling involved or with bonding tendencies. What seems to count is relevance of whatever people interact with in relation to the task at hand (but with eyes open, please);

5. Even if we tried very hard to accommodate the theory, no shimmer of a singularity experience was present in our data.

To the unsuspected user, putting a highly-acclaimed robot like Hanson's Sophia in remote control while being driven by human intelligence basically *is* the singularity:

Although the Singularity has many faces, its most important implication is this: our technology will match and then vastly exceed the refinement and suppleness of what we regard as the best of human traits. (Kurzweil, 2005, Chap. 1) [19]

Letting a human follow robot protocol favors Media Equation, CASA, and singularity assumptions: We provoked false-alarms-to-robots for the actual human and let those detection flaws coincide with missing-signals-to-robots for the actual robot, so that all would become equated or at least less different (Hoorn & Tuinhof, 2022) [11]. But that did not help. As an example, when our Cantonese human confederate kept to her English-spoken protocol (the one also followed by the robot operators), a Cantonese participant asked 'Why don't you just speak Cantonese to me?' Humans are expected to have default adaptability of affordances that robots are not expected to have.

Putting the human confederate at a disadvantage would make her come closer to the robots. What happened, however, was that we encountered a concept-driven bias: Knowing it was a human made people prefer the agency they worked with. As demonstrated empirically in this paper, natural intelligence limited by a robot protocol and by restrained electro-mechanics showed no different performance than a person keeping strict (robot) protocol. If people believe an avatar created in MetaHuman Creator's Unreal Engine is a real human being, they would equally prefer that avatar. Media Equation is anthropomorphism while being aware of the category mismatch. In our study, we saw concept-driven attribution of humanlike qualities to both machines *and* humans based on a perceived category match. If people believed the actual human was a machine, they would equally have disliked the human.

Therefore, we conclude for Media Inequality and a flat-topped Uncanny Mountain. After all, we could not establish a downfall in involvement among the three artificial agencies but only an insignificant increase in feeling distant. On the affective side of the user experience, the interaction felt somewhat uncanny, there was a pinch of Media Equation, and the singularity was far far away. On the functional side, people thought that 'You are no different from a robot but we like you better because you are human.' There were no effects of being human on task-related experiences so that a human strictly following protocol did not do better than machines; it's just that a human's human-likeness raises less distance and more involvement.

… as long as there is an AI shortcoming in any such area of endeavor, skeptics will point to that area as an inherent bastion of permanent human superiority over the capabilities of our own creations. (Kurzweil, 2016, p. 148) [20]

Indeed. For now, the singularity is near only if humans try to behave like their machines. We hope our study is an appeal to reason rather than proclaiming mind-blowing visions of media seen as real people, a robot check-in as a scary Zombie Hunt laser-tag battle at Hong Kong's Ocean Park, or of Singularity beyond biology. "I shall endeavor to function adequately, sir." Perhaps android Data should leave it just there.

## 5.1 Robot and digital character design

Tribute to robot Data's functional adequacy, the primary thing a designer should do is to gauge whether human-likeness is appreciated by the stakeholders of an interactive system such as social robots and avatars. Does impersonating a human being help task execution, general experience, or maybe the marketing aspects of the service? If so, which robot features should be humanlike? Is looking at the customer helpful in any way? What we do know is that a full-scale robot equivalent of a human being (i.e. attempting the singularity) is a deep investment that may overshoot the goal as we saw with Sophia robot.

Design focus nowadays is on human-realistic appearance such as a lifelike Frubber skin or realistic behaviors like keeping politeness rules but robots are not perceived as other human beings. Robots have their own rules of conduct, somewhat resembling those for real humans (case in point for CASA). If at all needed, our PCA results for human-likeness go beyond appearance and behavior and call attention to far deeper and more complicated matters such as having 'a soul' or 'free will,' which may not be programmable but can only be suggested, for instance, through story-telling, whether by the robot itself or via its personal website and social media. When an artist paints a portrait, s/he copies certain characteristics of the living subject to a non-living medium: canvas and paint. However moving that portrait may be, no one will take it for the real person. Be aware, then, that users do not want to be fooled so the fiction should remain clear, which brings the storytelling designer close to being a novelist.

One can keep an eye on uncanny effects but they are incidental and probably wear off quickly, particularly when someone is satisfied with what the machine should provide, for example, companionship or an efficient check-in service. If human-likeness is crucial for task execution, we should have found in our experiment that the human confederate 'did better' than the other agencies but she did not. If warmth and empathy are important to the task at hand, humans like humans better because they are humans (an in-group effect?) but those who provide service could follow a procedure just the same (cf. a therapist running a depression protocol). Often, managers instruct their employers as if they were a robot (cf. *robotomorphism* in Hoorn & Tuinhof, 2022 [11]).

Humans are highly complex biochemical open-state systems that label certain internal states and processes as mind, soul, intelligence, and consciousness. Robots are simpler, electro-mechanical closed-state machines that process sequences of subatomic particles without regarding that as information. People experience the effects of those differences and treat humans differently from robots. They do acknowledge, however, that certain features of humans can be designed into the non-living medium of steel and plastics. At times, then, they treat robots as if they were humans, quite like a child during doll play.

Most will agree that Mona Lisa moves observers more than the chairperson of the European Union. Certain artificial representations emotionally feel better than the real thing. If done right, people may prefer a robot as a partner over a human being. That does not mean, however, that they are unaware of the nonhuman nature of the doll they play with.

**ACKNOWLEDGMENTS**

This field experiment was a full-blown private-public international cooperation to which many contributed. We are grateful to the Hong Kong Productivity Council provided us with the Iwaa robot as well as organizational and technical support: We thank Samantha Chan, Ming Ge, Venus Wu, Crystal Cheung, and Suki Fan, Edison, and Eren. We thank Hong Kong Science Park, especially Henry So and Wendy Py Wan, for having our experiment take place at one of their events. Likewise, we thank Hanson Robotics Ltd for lending us their Sophia robot and Sophia avatar and provide us with (technical) support: We thank Amit Kumar Pandey, Luisa Zhou, Ralf Mayet, Mario Guzman, Jeanne Lim, and David Hanson himself. From Vrije Universiteit Amsterdam, we received on-site support from Stefanie de Boer, Dominique Roos,

Marijn Hagenaar, and Lars van Buuren in the organization of the experiments and collecting the data. Elly A. Konijn is kindly acknowledged for the supervision of the questionnaire design. Marsha Lui from The Hong Kong Polytechnic University was responsible for carefully translating the questionnaire items into Chinese/Cantonese. From that same university, Kathy Chan and Zhang Xinqyu are kindly thanked for their assistance during the experiments. We are thankful to Johnny K. W. Ho for his invaluable advice.